\documentclass{article}

\usepackage{spconf}
\usepackage{amsmath,amssymb,amsfonts}
\usepackage{fixmath}
\usepackage{graphicx}
\usepackage{algorithm}
\usepackage[noend]{algpseudocode}
\usepackage{threeparttable}
\usepackage{booktabs}
\usepackage{hyperref}
\graphicspath{{./figures/}}

\newcommand{\mb}{\boldsymbol}

\newcommand{\bx}{\mb{x}}

\newcommand{\by}{\mb{y}}

\newcommand{\bz}{\mb{z}}
\newcommand{\bZ}{\mb{Z}}

\newcommand{\bw}{\mb{w}}
\newcommand{\bv}{\mb{v}}

\newcommand{\bh}{\mb{h}}
\newcommand{\bR}{\mb{R}}

\newcommand{\bC}{\mb{C}}

\newcommand{\blam}{\mb{\lambda}}
\newcommand{\bLam}{\mb{\Lambda}}

\newcommand{\bmu}{\mb{\mu}}
\newcommand{\bSig}{\mb{\Sigma}}

\newcommand{\eps}{\epsilon}
\newcommand{\beps}{\mb{\eps}}

\newcommand{\bXi}{\mb{\Xi}}
\newcommand{\bxi}{\mb{\xi}}

\def\E{\mathbb{E}}

\newcommand{\nn}{\nonumber}

\title{NONLINEAR KALMAN FILTERING WITH REPARAMETRIZATION GRADIENTS}
\name{San Gultekin, Brendan Kitts, Aaron Flores, and John Paisley*}
\address{Yahoo! Inc., *Columbia University \\ \texttt{ \small \{sgultekin, bkitts, aaron.flores\}@yahooinc.com, jpaisley@columbia.edu } }

\begin{document}
\ninept

\maketitle

\begin{abstract}
We introduce a novel nonlinear Kalman filter that utilizes reparametrization gradients. The widely used parametric approximation is based on a jointly Gaussian assumption of the state-space model, which is in turn equivalent to minimizing an approximation to the Kullback-Leibler divergence. It is possible to obtain better approximations using the alpha divergence, but the resulting problem is substantially more complex. In this paper, we introduce an alternate formulation based on an energy function, which can be optimized instead of the alpha divergence. The optimization can be carried out using reparametrization gradients, a technique that has recently been utilized in a number of deep learning models.
\end{abstract}


\section{Introduction}
\label{sec:intro}

Nonlinear Kalman filtering is a useful tool for numerous signal processing and machine learning applications; several examples include target tracking \cite{LiJilkov1}, time series forecasting \cite{Gultekin_2019}, collaborative filtering \cite{Gultekin_2014}, and natural language processing \cite{Belanger_2015}. While the Kalman filter is optimal for the linear-Gaussian systems, the problems that fall under the umbrella of the nonlinear variant requires approximation, which can be divided into: (i) assumed density filtering, which is a parametric approximation \cite{Maybeck_1982}, and (ii) particle filtering, which is nonparametric and relies on Markov Chain Monte-Carlo \cite{Arulampalam_2002}.

A jointly Gaussian approximation of the state-space model is widely used in assumed density filtering. For many applications such as target tracking, Gaussian distribution is a natural modeling choice. In addition, a Gaussian random variable is completely characterized by its mean and variance, rendering the filtering problem computationally tractable. Widely used filters such as the Extended Kalman Filter (EKF), Unscented Kalman Filter (UKF) \cite{Julier_2004}, and its generalizations \cite{Guo_2006} are based on the aforementioned jointly Gaussian assumption. In our previous work \cite{Gultekin_2017}, we have investigated this assumption from the perspective of Bayesian inference, and showed that the filters mentioned above are all optimizing an approximation to the variational lower bound \cite{Gultekin_2017, Bishop_2006}. The variational lower bound, in turn, arises when one wants to minimize the Kullback-Leibler divergence between the true posterior distribution and an approximating distribution, which in this case is Gaussian. In \cite{Gultekin_2017}, we have introduced a novel filter which optimizes the exact variational lower bound, which results in improved performance.

While the variational lower bound arises from the KL divergence, changing the arguments in the function results in a different optimization problem, known as Expectation Propagation (EP) \cite{Minka_2001}. More generally, we can consider the family of alpha divergences, indexed by a single parameter. By changing the value of $\alpha$, we can obtain a class of filters. In \cite{Gultekin_2017}, we have tackled this more general problem using moment matching, akin to the Power EP \cite{Minka_2004}. However, similar to the latter, our moment matching is not guaranteed to converge either, although if it converges, it will be at a stationary point of the alpha divergence. To deal with the convergence issues of EP, alternative formulations based on energy functions from statistical physics were considered; in \cite{Opper_2005} the authors present an energy function formulation and a double-loop algorithm which is guaranteed to converge. For the alpha divergence objective of Power EP, an energy function was formulated in \cite{Minka_2004}.

In this paper we introduce a novel Kalman filter based on an energy function formulation. While these functions are typically difficult to optimize \cite{Minka_2004}, for the filtering problem we show that an elegant solution exists using reparametrization gradients, a technique that is gaining increasing traction in deep learning research; in particular the variational autoencoders \cite{Kingma_2014} are trained using these gradients. Generally speaking, in Bayesian statistics, approximate inference under different divergences is a field of active research, although the primary focus is not on dynamic models. Some examples include $\alpha$-divergence \cite{Lobato_2016}, $\chi$-divergence \cite{Dieng_2017}, Wasserstein divergence \cite{Ambrogioni_2018}, and Renyi divergence \cite{iturner_2016}. In particular, there are two papers that use an energy function related to ours. These are the ones that use the alpha divergence \cite{Lobato_2016} and Renyi divergence \cite{iturner_2016} for batch inference problems. Both these papers and ours use energy functions that can be viewed as descendants of the Power EP energy function \cite{Minka_2004}. With that said, however, our approach is distinct from those papers in the following ways: (i) In \cite{Lobato_2016} and \cite{iturner_2016}, the objective function for the batch case is obtained by simplification, whereas our function is exact. (ii) The choice of prior is free in batch models, whereas in our case it is dictated by the dynamic model. (iii) Our optimization algorithm is different from the two aforementioned papers; \cite{Lobato_2016} applies reparametrization directly whereas \cite{iturner_2016} uses importance reweighing. While their strategy is effective for the batch case, for our scenario the gradients produced either way can become numerically unstable. On the other hand, we introduce a scaling trick that gives stable gradients. (iv) Finally, we show how to obtain natural gradients from the reparametrized gradients, which can be used for full covariance estimation, a difficult problem often ignored in stochastic gradient descent.

\section{Methodology}
\label{sec:method}

The starting point of the nonlinear Kalman filtering problem is the following state-space model
\begin{align}\label{eq:ss_model}
	\bx_t &= f(\bx_{t-1}, \bw_t) \nn \\
	\by_t &= h(\bx_t, \bv_t)
\end{align}
where $f(\cdot)$, $h(\cdot)$ are nonlinear functions, and $\bw_t$, $\bv_t$ are noise terms. Here $\{\bx_t\}_{t=1}^T$ are the set of latent variables that are to be estimated, using the observations $\{\by_t\}_{t=1}^T$. In particular, we are interested in the filtering problem, which causally estimates $\bx_t$'s given observations up to the current time. The posterior distribution of interest is then $p(\bx_{1:t} | \by_{1:t})$. By Markov property and Bayes' rule it is straightforward to show that $p(\bx_t | \by_{1:t}) \propto p(\by_t | \bx_t) ~ p(\bx_t | \by_{1:t-1})$, which indicates that the posterior can be calculated recursively. In the first step we calculate the prior, namely $p(\bx_t | \by_{1:t-1})$, from $p(\bx_{t-1} | \by_{1:t-1})$; in the second step we use this prior, along with the likelihood term, to obtain the posterior. For linear-Gaussian systems, both the prior and posterior distributions are Gaussian. In Bayesian statistics this is known as conjugate distributions -- given the likelihood, the prior and posterior distributions have the same parametric form.

Practically speaking, for the more general model in Eq. \eqref{eq:ss_model}, conjugacy is no longer guaranteed. In this case, an approximation is required. One widely used method is variational inference \cite{Jordan_1999}, where the intractable posterior distribution is approximated by a tractable one, which we refer to as the $q$-distribution. The quality of approximation is measured by Kullback-Leibler divergence, which is written as
\begin{align}
	\text{KL}[q(\bx)||p(\bx)] = \int q(\bx) \log \frac{q(\bx)}{p(\bx)} d\bx ~.
\end{align}
For our problem, the $p$-distribution and $q$-distribution are the true posterior and approximating normal at time $t$, namely $p(\bx_t | \by_t)$ and $q(\bx_t)$. The KL divergence is equal to zero when $p(\bx) = q(\bx)$. With that said, it is not possible to directly minimize the KL divergence, as it requires the knowledge of true posterior. In variational inference, we instead maximize the lower bound
\begin{align}
	\mathcal{L}[q(\bx)] = \int q(\bx) \log p(\bx, \by) d\bx - \int q(\bx) \log q(\bx) d\bx ~.
\end{align}
Maximizing this objective is equivalent to minimizing the KL divergence, since $p(\by) = KL[q(\bx)||p(\bx)] + \mathcal{L}[q(\bx)]$. In \cite{Gultekin_2017}, we have proven that filters such as EKF and UKF are maximizing an approximation to this lower bound. In that paper, we have also shown how one can instead obtain unbiased gradients of $\mathcal{L}[q(\bx)]$ based on the parameters of the $q$-distribution. In the case of normal distributions, this amounts to calculating gradients with respect to the mean and covariance, which we denote $\nabla_{\bmu} \mathcal{L}$ and $\nabla_{\bSig} \mathcal{L}$ respectively. We have also shown how natural gradients could be utilized for numerical stability.

In this paper, we are interested in utilizing the aforementioned gradient method for alpha divergence minimization. The alpha divergence is defined by
\begin{align}\label{eq:alpha_div}
	D_{\alpha}[p(\bx)||q(\bx)] = \frac{1}{\alpha(1-\alpha)} \left( 1 - \int p(\bx)^{\alpha} q(\bx)^{1-\alpha} d\bx \right) ~.
\end{align}
Alpha divergence defines a family of divergences that include the two KL divergences as special cases, in particular, $D_{\alpha}[p||q] \rightarrow KL[q||p]$ as $\alpha \rightarrow 0$, and $D_{\alpha}[p||q] \rightarrow KL[p||q]$ as $\alpha \rightarrow 1$. In \cite{Gultekin_2017}, we proposed an importance sampling procedure similar in spirit to the Power EP. However, both methods have a drawback in that, while they converge to a stationary point of the objective function, convergence is not guaranteed. Naturally, this leads to the question of designing a procedure that converges. The answer to the above problem is given by the energy functions.

\subsection{Energy Functions}

Recall that a random variable $\bZ$ has an exponential family distribution if it can be written as $p(\bz) = h(\bz) \exp\{s(\bz)^\top \blam_z - \log Z(\blam_z)\}$ where $h(\bz)$ is the base measure, $s(\bz)$ is the sufficient statistics, $\blam_z$ is the natural parameter, and $\log Z(\cdot)$ is the normalizing function \cite{Nielsen_2014}. For example, for the Gaussian distribution, the canonical parameters are the mean and covariance, which we denote $(\bmu, \bSig)$, while the natural parameters are $(\bLam \bmu, \bLam)$, where $\bLam = \bSig^{-1}$ is the precision matrix. Since we are interested in Gaussian approximations, the prior distribution will be in exponential family form, which we write as
\begin{align}\label{eq:prior_expfam}
	p(\bx_t | \by_{1:t-1}) = h(\bx_t) \exp \{ \blam_{t|t-1}^\top s(\bx_t) - \log Z(\blam_{t|t-1}) \} ~.
\end{align}
We are now interested in finding an approximating distribution, such as $p(\bx_t | \by_t) \approx q(\bx_t)$. The objective is to minimize the alpha divergence, which can be cast as $\min_{q(\bx_t)} D_{\alpha}[p(\bx_t | \by_t) || q(\bx_t)]$. Now, let the approximating distribution be in the same exponential family, i.e. $q(\bx) = h(\bx) \exp \{ \blam_q^\top s(\bx) - \log Z(\blam_q) \}$. The derivative of the alpha divergence with respect to $\blam_q$ is
\begin{align}
	\nabla_{\blam_q} D_{\alpha} [p||q] = - \frac{Z_{\alpha}}{\alpha} \int \frac{p(\bx)^\alpha q(\bx)^{1-\alpha}}{Z_{\alpha}} \bigg[ s(\bx) - \E_q [s(\bx)] \bigg]
\end{align}
where the exponential family identity $\nabla_{\blam_q} \log Z(\blam_q) = \E_q[s(\bx)]$ is used. Above, $Z_{\alpha}$ is the normalizing constant of the distribution $\tilde{p}(\bx) \propto p(\bx)^\alpha q(\bx)^{1-\alpha}$. Setting the derivative to zero, the distribution that minimizes the alpha divergence satisfies
\begin{align}
	\E_{\tilde{p}(\bx)} [s(\bx)] = \E_{q(\bx)} [s(\bx)]
\end{align}
which is a generalized moment matching condition \cite{Gultekin_2017}.

In this paper, we are interested in an alternate formulation which would make alpha divergence optimization viable. For the following discussion we drop the time indices for clarity. Let the prior distribution be given by $p_0(\bx) = \exp\{\blam_0^\top s(\bx) - Z(\blam_0)\}$ and the approximating distribution be from the same exponential family $q(\bx) = \exp\{\blam_q^\top s(\bx) - Z(\blam_q)\}$. Also define $\blam = \blam_q - \blam_0$, which is known as the ``cavity parameter " \cite{Minka_2001}. Based on this, define $f(\bx) = \exp\{\blam^\top s(\bx)\}$. The energy function is given by
\begin{align}\label{eq:energy}
	E(\blam_q) = \log Z(\blam_0) - \log Z(\blam_q) - \frac{1}{\alpha} \log \E_q \left[ \frac{p(\by|\bx)^\alpha}{f(\bx)^\alpha}\right] ~.
\end{align}

Next, we show that the stationary points of this energy function is the same as that of the alpha divergence.
\begin{align*}
	\nabla_{\blam_q} E(\blam_q) = -\E_q[s(\bx)] - \frac{1}{\alpha} \frac{\nabla_{\blam_q} \int q(\bx) p(\by|\bx)^\alpha f(\bx)^{-\alpha} d\bx}{\int q(\bx) p(\by|\bx)^\alpha f(\bx)^{-\alpha} d\bx}
\end{align*}
where we once again use $\nabla_{\blam_q} Z(\blam_q) = \E_q[s(\bx)]$. The denominator of the latter term evaluates as
\begin{align*}
	\int q(\bx) p(\by|\bx)^\alpha f(\bx)^{-\alpha} d\bx = Z(\blam_0)^\alpha Z(\blam_q)^{-\alpha} p(\by)^\alpha Z_{\tilde{p}} ~.
\end{align*}
Next define $\Omega(\blam_q) = \exp \{ (1-\alpha) \blam_q^\top s(\bx) - \log Z(\blam_q) \}$. Then
\begin{align*}
	\nabla_{\blam_q} E(\blam_q) &= -\E_q[s(\bx)] - \frac{1}{\alpha} \frac{\nabla_{\blam_q} \int p(\bx|\by)^\alpha \Omega(\blam_q)}{Z_{\tilde{p}} Z(\blam_q)^{-\alpha}} ~ dx \\
	&= -\E_q[s(\bx)] - \frac{1}{\alpha} \frac{\int p(\bx|\by)^\alpha \nabla_{\blam_q} \Omega(\blam_q)}{Z_{\tilde{p}} Z(\blam_q)^{-\alpha}} ~ dx \\
	&= -\E_q[s(\bx)] - \frac{1}{\alpha} \int \frac{p(\bx|\by)^\alpha q(\bx)^{1-\alpha}}{Z_{\tilde{p}}} ~ \times \\
	&\quad\quad\quad\quad [(1-\alpha) s(\bx) - \E_q[s(\bx)]] ~ d\bx \nn \\
	&= - (1 - 1/\alpha) [\E_q[s(\bx)] - \E_{\tilde{p}}[s(\bx)]] ~,
\end{align*}
which shows that the stationary points are the same as $D_\alpha[p||q]$. Therefore, we can work directly with the energy function in Eq. \eqref{eq:energy}.

\subsection{Multivariate Normal}

The multivariate normal is characterized by the mean $\bmu$, and covariance $\bSig$ (or equivalently precision $\bLam = \bSig^{-1}$), and has the functional form
\begin{align}
p(\bx) = (2\pi)^{-d/2} \log |\bSig|^{-1/2} \exp \{ -\frac{1}{2} ~ (\bx-\bmu)^\top \bSig^{-1} (\bx-\bmu) \} ~.
\end{align}
In exponential family form this can be re-written as
\begin{align}
	p(\bx) &= (2\pi)^{-d/2} \exp \{-1/2 ~ \text{vec}(\bLam)^\top \text{vec}(\bx \bx^\top) \nn \\
	&\quad + (\bLam \bmu)^\top \bx -1/2 ~ \bmu^\top \bLam \bmu + 1/2 ~ \log |\bLam| \}
\end{align}
where $\blam = [\bLam^\top\bmu ~,~ \text{vec}(\bLam)]$\footnote{~$\text{vec}(\cdot)$ flattens a given matrix into a vector.} is the natural parameter, $s(\bx) = [\bx ~,~ \text{vec}(\bx\bx^\top)]$ is sufficient statistics, and $Z(\blam) = -1/2 ~ \bmu^\top \bLam \bmu + 1/2 ~ \log |\bLam|$.

For the model we consider in nonlinear Kalman filtering, the prior is $p_0(\bx_t) \sim N(\bmu_0, \bSig_0)$. The latent variable is linked to the observation via the conditional mean, namely $p(\by_t|\bx_t) \sim N(h(\bx_t),\bSig)$ where $h(\cdot)$ is a known arbitrary first-order differentiable function, and $\bSig$ is a known constant. In general, the posterior distribution for this case is not Gaussian; however, to facilitate conjugacy, we pick the approximating distribution to be Gaussian, such that $q(\bx_t) \sim N(\bmu_q, \bSig_q)$ with slight abuse of notation.

Due to the 1-1 correspondence between the natural and mean parameters, the partition functions in Eq. \eqref{eq:energy} can be written as
\begin{align}
    \log Z(\bmu_q, \bSig_q) = \frac{1}{2} \bmu_q^\top \bSig_q^{-1} \bmu_q + \frac{1}{2} \log |\bSig_q| ~,
\end{align}
which can be differentiated directly with respect to $\bmu_q$ and $\bSig_q$. The remaining term is a bit complicated as it depends on $\bx$ and $h(\cdot)$. Ignoring the constant terms and time indices, we have
\begin{align} \label{eq:nonseparable}
    \log \E_q \biggl[ (\by - \bh)^\top \bR^{-1} (\by - \bh) + \bx^\top \bXi \bx + 2\bx^\top \bxi \biggr] ~,
\end{align}
where $\bh = \bh(\bx)$, $\bXi = \bSig_q^{-1} - \bSig_0^{-1}$, and $\bxi = \bSig_0^{-1}(\bmu_0 - \bmu_q)$. Note that, the difficulty here is due to the expectation itself depends on the parameters we want to differentiate. At this point, we utilize the reparametrization trick which elegantly resolves the issue. Let $\beps_s \sim N(0, \mb{I})$ and $\bC$ be the Cholesky decomposition of $\bSig_q$. Then we can generate a sample $\bx_s$ from $q(\bx)$ using the transform $\bx_s = \bC \beps_s + \bmu_q$. Using this sample, Eq. \eqref{eq:nonseparable} can be written as
\begin{align}
    \log \E_q \biggl[ (\by - \bh_s)^\top \bR^{-1} (\by - \bh_s) + \bx_s^\top \bXi \bx_s + 2 \bx_s^\top \bxi \biggr].
\end{align}
where $\bh_s = \bh(\bx_s)$. In practice, we use $S$ samples to obtain a less noisy estimate of the energy function. The final objective function is shown in Line 9 of Algorithm 1. Furthermore, we can compute the gradients automatically, using automatic differentiation. For example, in our implementation, we utilize the Autograd package for Python, which is available for free. The link to our code is given in Section \ref{sec:exp}.

We now describe a specific gradient computation that is suitable for covariance matrices. In particular let $\nabla_{\bmu_q} E(\bmu_q^i, \bSig_q^i)$ and $\nabla_{\bSig_q} E(\bmu_q^i, \bSig_q^i)$ be the gradients computed at $i$-th iteration. We perform the following updates
\begin{align}
	\bmu_q^{i+1} &\gets \bmu_q^i - \rho_i \bC_{\bSig}^i \nabla_{\bmu_q} E(\bmu_q^i,\bSig_q^i) ~, \nn \\ \bSig_q^{i+1} &\gets \bSig_q^i - \rho_i \bC_{\bSig}^i \nabla_{\bSig_q} E(\bmu_q^i,\bSig_q^i) \bC_{\bSig}^i
\end{align}
where $\bC_{\bSig}^i = \bSig_q^i$ is a conditioning matrix and $\rho_i$ is the step size. This update is linked to the natural gradient of the parameters \cite{Yi_2009} and is particularly suitable for gradient descent with full covariance matrices, as it takes the curvature of the space into account. The final algorithm is summarized in Algorithm 1.

\begin{algorithm}[t]
	\caption{Energy Function based Kalman Filter (EFKF\textsubscript{$\alpha$})}
	\textbf{Input:} Prior: $p_0(\bx_t) \sim N(\bmu_0, \bSig_0)$ \\
	\phantom{\textbf{Input:}} Likelihood: $p(\by_t|\bx_t) \sim N(h(\bx_t), \bSig)$ \\
	\textbf{Output:} Posterior: $q(\bx_t) \sim N(\bmu_q, \bSig_q)$ \\
	1.~ Initialize: $\bmu_q^0 \gets \bmu_0$ and $\bSig_q^0 \gets \bSig_0$ \\
	2.~ \textbf{For} $i \in {0,\ldots,I-1}$ \\
	3.~ \quad {\tt //Approximate energy function} \\
	4.~ \quad $\forall s:$ $\beps_s \sim N(0, \mb{I})$ \\
	5.~ \quad $\forall s:$ $\bx_s \gets \bC \beps_s + \bmu_q^i$ \hfill ($\bC \gets \text{chol}\{\bSig_q^i\}$) \\
	6.~ \quad $\forall s:$ $\Psi(s) \gets \alpha \log p(\by|h(\bx_s), \bSig) - \alpha \log f(\bx_s)$ \\
	7.~ \quad $\Psi \gets \max\{\Psi(1),\ldots,\Psi(S)\}$ \\
	8.~ \quad $\forall s:$ $\hat{\Psi}(s) \gets \Psi(s) - \Psi$ \\
	9.~ \quad $\hat{E}(\bmu_q^i,\bSig_q^i) = \log Z(\bmu_0,\bSig_0) - \log Z(\bmu_q^i,\bSig_q^i) \\
	\phantom{9.~ \quad} \quad -\frac{1}{\alpha}\log \frac{1}{S} \sum_{s=1}^S e^{\hat{\Psi}(s)} - \frac{1}{\alpha} \Psi$ \\
	10. \quad {\tt //Gradient update} ($\bC_{\bSig}^i \gets \bSig_q^i$) \\
	11. \quad $\bmu_q^{i+1} \gets \bmu_q^i - \rho_i \bC_{\bSig}^i \nabla_{\bmu_q} \hat{E}(\bmu_q^i, \bSig_q^i)$\\
	12. \quad $\bSig_q^{i+1} \gets \bSig_q^i - \rho_i \bC_{\bSig}^i \nabla_{\bSig_q} \hat{E}(\bmu_q^i, \bSig_q^i) \bC_{\bSig}^i$ \\
	13. \textbf{EndFor} \\
\end{algorithm}

\section{Numerical Results}
\label{sec:exp}

\begin{table}[t]
	\caption{Root mean square errors (RMSE) for the tracking problem. We can see that EFKF (EF) provides robust estimates for the parameter mismatch scenario, achieving the best results with $\alpha=0.7$. More interestingly, this configuration also yields the best performance for matching parameters, overperforming the particle filter.}
	\label{tab1}
	\centering
	\begin{tabular}{lcccc|c}
		\toprule
		& & Mismatch & & & Match \\
		\cmidrule{2-6}
		$$ & $0.01\times$\textbf{I} & $0.05\times$\textbf{I} & $0.1\times$\textbf{I} & $0.5\times$\textbf{I} & $\text{Q}_\text{CV}$ \\
		\midrule
		$\text{EF}_{.01}$ & 12.7497 & 11.3567 & 10.8337 & 12.0719 & 12.7419 \\
		$\text{EF}_{.10}$ & 12.4064 & 11.0164 & 10.6125 & 11.1629 & 12.4490 \\
		$\text{EF}_{.30}$ & 11.9321 & 10.3581 & 10.2275 & 10.9492 & 12.0167 \\
		$\text{EF}_{.50}$ & 10.9544 &  9.7896 &  9.9947 & 10.8950 & 10.9482 \\
		$\text{EF}_{.70}$ &  9.9618 &  \textbf{9.6794} &  9.8298 & 11.0407 &  \textbf{9.6120} \\
        $\text{SKF}$ & 12.7126 & 11.4448 & 10.8847 & 11.5156 & 12.7441 \\		
        $\text{MKF}$ & 12.7076 & 11.0775 & 10.4697 & 10.9253 & 12.6978 \\
		\midrule
		$\text{EKF}$ & 16.4635 & 17.0646 & 17.5063 & 17.7530 & 16.6406 \\
		$\text{UKF}$ & 13.8852 & 12.7018 & 11.9827 & 11.8454 & 13.8841 \\
		$\text{PF}$ & 11.6420 & 10.9945 & 10.8421 & 11.0859 & 11.6019 \\
		\bottomrule
	\end{tabular}
\end{table}

\begin{figure*}[t]
	\centering
	\includegraphics[width=\textwidth]{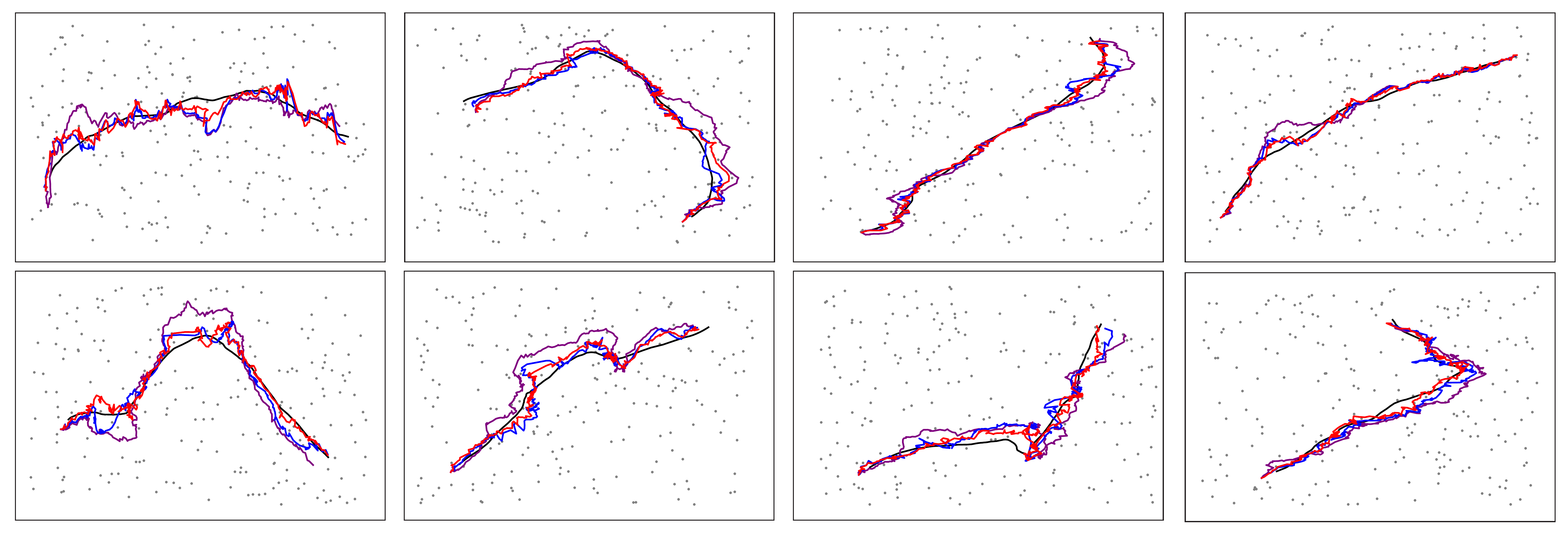}
	\caption{Eight sample paths from the tracking experiment. Original trajectory (black), EKF (purple), PF (blue), and $\text{DEM}_{0.70}$ (red) are shown. The sensors are shown as grey dots. Qualitatively, it can be seen that DEM provides the most accurate estimates of the target state.}
	\label{fg:paths}
\end{figure*}

In this section we illustrate and benchmark EFKF on a tracking experiment which is often used to compare various nonlinear filters \cite{Gultekin_2017}. The code for experiments is available online.\footnote{ \url{https://github.com/sangultekin/nonkf_energy_minimization}}

The object moves in a 2D space based on the following constant-velocity model
\begin{align}
	\bx_t = \mb{F}_t \bx_{t-1} + \mb{w}_t \quad,\quad \mb{w}_t \sim N(\mb{0}, \mb{Q}_t)
\end{align}
where
\begin{align}
	\mb{F_t} &= \begin{bmatrix} \mb{F}_2 & \mb{0} \\ \mb{0} & \mb{F}_2 \end{bmatrix} ~,~ \mb{F}_2 = \begin{bmatrix} 1 & \Delta_t \\ 0 & 1 \end{bmatrix} ~,~ \nn \\
	\mb{Q}_t &= \begin{bmatrix} \mb{Q}_2 & \mb{0} \\ \mb{0} & \mb{Q}_2 \end{bmatrix} ~,~ \mb{Q}_2 = \sigma_{CV} \begin{bmatrix} \Delta_t^4/4 & \Delta_t^3/2 \\ \Delta_t^3/2 & \Delta_t^2 \end{bmatrix} ~.
\end{align}
This model is capable of generating a wide variety of sample paths, as illustrated in Figure \ref{fg:paths} (original paths are in black). The latent state of the object is then the position and velocity of the object, and the measurements are given by sensors distributed across the field.

For our experiments, at each time three sensors are active and $h(\bx_t)$ is a three dimensional vector where $[h(\bx_t)]_i = \|\bx_t-\mb{s}_{t,i}\|$ is the distance to $i$-th sensor. The likelihood function is then given by $p(\by_t | \bx_t) = N(\by_t | h(\bx_t), \mb{R})$ where $\mb{R}$ is a known matrix, based on sensor characteristics. For this model we shall choose the approximating posterior Gaussian as well. This is similar to the nonlinear Kalman filters which make Gaussian approximation to the posterior using various methods. In particular we will compare our method against the following: (i) Extended Kalman Filter (EKF) where $h(\bx_t)$ is approximated by a first order Taylor expansion, (ii) Unscented Kalman Filter (UKF) which uses sigma-points to approximate $h(\bx_t)$, (iii) Particle Filter (PF) which is the golden standard nonparametric filter for nonlinear state-space models \cite{Arulampalam_2002}, (iv) Ensemble Kalman Filter (ENKF) which uses data assimilation \cite{Evensen_2003}, and (v)
Stochastic Search Kalman Filter (SKF) which uses stochastic search method to optimize variational lower bound \cite{Gultekin_2017}. For our filter based on energy minimization (EFKF) we experiment with five different values as shown in the leftmost column of Table 1. In particular we note that EFKF with $\alpha=1$ corresponds to matching moments directly instead of performing gradient descent.

Table \ref{tab1} shows quantitative results for the tracking experiment. We consider two different scenarios. In the first one the covariance of the process noise ($\mb{Q}_{CV}$) is known with some uncertainty, and we need to sweep across a range of covariance estimates (chosen as four here due to space limitations) as shown under the ``Mismatch" columns. The values are reported as the root mean square error and averaged over one hundred runs to ensure statistical significance of the results. Due to width constraints we do not show the errors in the Table. It can be seen that overall the EKF approximation is poor and has the worst performance. UKF and ENKF improve upon this, where UKF is more robust. However PF outperforms all three of them, as expected. As for EFKF, for low and high values of $\alpha$ the results are comparable to PF where EFKF performs better in some cases and PF better in the others. The same also applies to SKF. On the other hand, for mid range of $\alpha$, EFKF is significantly better than all the other filters, including PF. The best value is highlighted in boldface, which occurs at $\alpha = 0.7$. These results corroborate the previous observation in \cite{Gultekin_2017} that, parametric filters offer a more robust alternative to the particle filter in case of parameter uncertainty. Interesting, unlike the previous case, we see that EFKF also provides better performance even when there is no parameter uncertainty. For this ``Match" we have the values in the rightmost column. Firstly, we can see that the PF outperforms all competitors including SKF and MKF, except for EFKF. For the case of $\alpha = 0.5$ and $\alpha = 0.7$, EFKF gives significantly lower RMSE. The best is once again highlighted in boldface and occurs at $\alpha = 0.7$.

We now show sample paths reconstructed from competing filters, for qualitative evaluation. In Figure \ref{fg:paths} w show eight sample paths; the estimates of location are provided by EKF, PF, and $\text{EFKF}_{0.70}$ for the scenario where there is no parameter mismatch. As the plots suggest, EKF has the highest susceptibility to noise in the measurement, and the estimates can deviate significantly at certain time points. PF improves significantly upon this, and produces more stable estimates. $\text{EFKF}_{0.70}$ further improves, and gives the best overall estimates, which was also reflected in Table \ref{tab1}.

\section{Concluding Remarks}
\label{sec:con}

We have introduced a novel nonlinear Kalman filter using energy minimization, which is inspired by approximate Bayesian inference using energy functions. The introduced filter directly optimizes the alpha divergence, addressing the convergence issues noted in previous literature. Experiments have shown significant performance improvements over previous divergence frameworks as well as particle filters. Furthermore, as hinted by Eq. \eqref{eq:energy}, our method is applicable to the more general case of exponential families, including but not limited to the case of Gaussian. In the full version of the paper we will show how the more general problem can also be addressed.

\bibliographystyle{IEEEbib}
\bibliography{references}

\end{document}